\documentclass[conference]{IEEEtran}
\IEEEoverridecommandlockouts
\usepackage{cite}
\usepackage{amsmath,amssymb,amsfonts}
\usepackage{algorithmic}
\usepackage{graphicx}
\usepackage{textcomp}
\usepackage{xcolor}
\def\BibTeX{{\rm B\kern-.05em{\sc i\kern-.025em b}\kern-.08em
    T\kern-.1667em\lower.7ex\hbox{E}\kern-.125emX}}
\begin{document}

\title{Graph Based Temporal Aggregation for Video Retrieval\\}

\author{
\IEEEauthorblockN{Arvind Srinivasan}
\IEEEauthorblockA{\textit{Dept. of CSE } \\
\textit{PES University}\\
Bangalore, India \\
arvind.srini.8@gmail.com}
\and
\IEEEauthorblockN{Aprameya Bharadwaj}
\IEEEauthorblockA{\textit{Dept. of CSE } \\
\textit{PES University}\\
Bangalore, India \\
aprameya.bharadwaj@gmail.com}
\and
\IEEEauthorblockN{Aveek Saha}
\IEEEauthorblockA{\textit{Dept. of CSE } \\
\textit{PES University}\\
Bangalore, India \\
aveek.s98@gmail.com}
\and
\IEEEauthorblockN{S Natarajan}
\IEEEauthorblockA{\textit{Professor, Dept. of CSE } \\
\textit{PES University}\\
Bangalore, India \\
natarajan@pes.edu}

}

\maketitle

\begin{abstract}
Large scale video retrieval is a field of study with a lot of ongoing research. Most of the work in the field is on video retrieval through text queries using techniques such as VSE++. However, there is little research done on video retrieval through image queries, and the work that has been done in this field either uses image queries from within the video dataset or iterates through videos frame by frame. These approaches are not generalized for queries from outside the dataset and do not scale well for large video datasets. To overcome these issues, we propose a new approach for video retrieval through image queries where an undirected graph is constructed from the combined set of frames from all videos to be searched. The node features of this graph are used in the task of video retrieval. Experimentation is done on the MSR-VTT dataset by using query images from outside the dataset. To evaluate this novel approach P@5, P@10 and P@20 metrics are calculated. Two different ResNet models namely, ResNet-152 and ResNet-50 are used in this study.
\end{abstract}


\section{Introduction}
Video Retrieval is one of the most eminent and challenging problems in the digital world today. It is the task of ranking videos in a database based on their relevance to user input queries. While most practical applications use video meta-data to convert the problem into a straight-forward page ranking problem, there are vast databases of videos with no labeled meta-data.

User queries can be of multiple forms. The most popular form of input is textual input. One of the most popular publicly available applications of text-to-video retrieval is Google’s video search. It’s performance, however, is driven by the metadata of the videos in its search space. Some research has been published in recent years on video retrieval through text queries and and the performance in this task has improved over the years.

In this paper, we focus our efforts on video retrieval through image queries, where an even lesser degree of research has been done. There are also drawbacks in most methods proposed so far and there are many challenges that any proposed method needs to overcome. The biggest of those challenges is to find an efficient method of storing the videos in your search database.

The next big challenge that hasn’t been solved yet is to tap into the temporal information contained in video data. While image classification and object detection tasks in images have been proven to work really well in the last few years, simply searching for objects in the frames of a video would fail to make use of the temporal information contained in videos.

In this paper, we propose a novel approach to solve this problem. We first pre-process a video database to extract key features from video frames. These features are clustered, such that similar frames end up in the same cluster. We generate the embeddings for these clusters by aggregating the embeddings of their constituent frames. To account for temporal information in videos, we model these clusters as nodes in a graph. Then, the cluster embeddings are augmented by including neighboring cluster information. These augmented cluster embeddings are stored and used in our video ranking process.

The proposed method was tested on the MSR-VTT dataset. 

\section{Related Work}

Most of the related work focuses on Video Retrieval through Text queries. Video retrieval
through image queries has been relatively unexplored. However, our approach borrows concepts
and other strategies like testing strategies from the discussed methods to develop a new approach
for video retrieval through image queries.

The paper \cite{araujo} introduces a new retrieval architecture, in which the image query can be compared
directly with database videos - significantly improving retrieval scalability compared with a
baseline system that searches the database on a video frame level.
Matching an image to a video is an inherently asymmetric problem. An asymmetric comparison
technique for Fisher vectors and systematically explore query or database items with varying
amounts of clutter, showing the benefits of the proposed technique. Novel video descriptors
which use Fisher vectors that can be compared directly with image descriptors are also proposed
in this work. Large-scale experiments using three datasets show that this technique enables faster and more
memory-efficient retrieval, compared with a frame-based method, with similar accuracy. The
proposed techniques are further compared against pre-trained convolutional neural network
features, outperforming them on three datasets by a substantial margin.
However, this paper only uses query images which are the frames of the videos in the dataset and does not work with out of dataset images.

The paper \cite{li} introduces a method of integrating the spacial temporal neighbourhood information using an attention mechanism that focuses on useful features on each frame. The Neighborhood Preserving Hashing method creates a learned hashing function that can easily map similar videos. 

Another approach for representation learning for videos is to create hierarchical graph clusters built upon video-to-video similarities. This is explored in the paper \cite{lee} using two different methods, the first is to create smart triplets and the second is to create pseudo labels. This creates a highly scalable  method for creating embeddings for video understanding tasks.

When videos are represented as individual frames it makes the modeling of long-range semantic dependencies difficult. The paper \cite{shao} solves this issue by incorporating long range temporal features at the frame level using self attention. For training on video retrieval datsets they propose a supervised contrastive learning method that performs automatic hard negative mining and utilizes the memory bank mechanism to increase the capacity of negative samples.

The paper \cite{dong} tackles the challenging problem of zero example video retrieval. In such a retrieval
paradigm, an end user searches for unlabeled videos by ad-hoc queries described in natural
language text with no visual example provided. Given videos as sequences of frames and queries
as sequences of words, an effective sequence-to-sequence cross-modal matching is required. The majority of existing methods are concept based, extracting relevant concepts from queries
and videos and accordingly establishing associations between the two modalities. In contrast, this
paper takes a concept-free approach, proposing a dual deep encoding network that encodes videos
and queries into powerful dense representations of their own. Dual encoding is conceptually
simple, practically effective and end-to-end. As experiments on three benchmarks, i.e. MSRVTT, TRECVID 2016 and 2017 Ad-hoc Video
Search show, the proposed solution establishes a new state-of-the-art for zero-example video
retrieval.

While most publications in the domain focus on learning joint text-video embeddings, there are
difficulties with the approach. There aren’t many caption-labeled video datasets to work with.
Hence, these methods aren’t too wide spread. To workaround this problem, they \cite{miech} use
heterogeneous data sources to learn text-video embeddings. They propose a new model called
Mixture-of-Embedding-Experts (MEE). It claims to work even with incomplete training data.
They extend their embedding technique to work with face descriptors. They evaluate their performance on the MPII Movie Description dataset and MSR-VTT dataset.
The proposed model shows considerable improvements and beats previous text-to-video retrieval
and video-to-video retrieval methods.

In the paper \cite{hu}, they introduce a semantic-based video retrieval framework. They use the
approach for surveillance videos. They detect motion trajectories using clustering based methods. These clusters are structured
hierarchically to obtain activity models. They propose a hierarchical structure of semantic
indexing and object retrieval. Here, each individual activity gets all the semantic descriptions of
the activity model from its parent activity. They use this technique to access individual objects
semantically. They also allow for different kinds of input queries like keyword search, queries by
sketch, and object queries.

This paper \cite{zhang} proposes an efficient method for video retrieval using image queries. The focus of
this approach is to make the entire process more efficient. They do not focus on improving the
accuracies of state-of-the-art models. Instead, they focus on building a model that will work even
for very large datasets. They use a Convolutional Neural Network to extract features from images in the videos. Then,
they use a Bag of Visual Words model to aggregate these features. They use the K-means
clustering algorithm to cluster them in order to reduce the space required to store the embeddings.
Then, they propose their Visual Weighted Inverted Index algorithm to improve the accuracy and
efficiency of retrieval. They evaluate their approach on the Youtube-8M and Sports-1M datasets. They use large datasets
to show that their model works well even with a large database of videos. They also compare the
performances of CNN-VWII and SIFT-VWII.

K-NN \cite{knn} is an unsupervised machine learning algorithm to cluster data points. Euclidean distance is used as the comparison metric. Euclidean distances of the incoming point with the centres of all the clusters are calculated. The shortest distance is found and the incoming data point is assigned to that particular cluster. If there are a lot of features in a data point, dimensionality reduction can be done before clustering. 

GraphSAGE \cite{hamilton} is an algorithm for inductive learning and representation on large graphs. It
generates low dimensional vectors for the nodes of the graphs. Existing models before this had to be
re-trained when a new node was added to the graph. GraphSAGE uses node information and neighbour information and aggregates them to generalize the features of the unseen node. 

Aggregators take the neighbourhood as input and combine the embeddings with certain weights to create embeddings for the neighbourhood. The initial embeddings of each node is set to its node features. Till the ‘K’ neighbourhood depth, the neighbourhood embedding is created using the aggregator function for each node and it’s
concatenated with the node features. This is then passed through a neural network to update the weights and features.

Residual networks are a class of deep neural networks proposed in \cite{szegedy}. If ‘x’ is the input to the initial layer H(x) is the mapping of ‘x’ for the first few layers of any deep network. In the case of the residual network, the mapping is modified into H(x) – x. We consider this as F(x). Therefore, H(x) = F(x) +
x. 

This formulation is used to solve the degradation problem. According to this problem, in theory
deeper networks should have the same training error as the shallow ones, but this problem shows
that due to the inability to approximate identity mappings by many non-linear layers. However,
with residual learning, the solvers make the weights of the non-linear layers almost zero to
approach identity mappings.
This learning is adopted to every few layers in the network. 

The input and output layers are x and y for a given layer and the function F(x, {Wi})
determines the residual mappings learned. This function also represents convolutional layers and
element wise addition is performed on the feature maps.
\section{Proposed Method}

The database of videos is first pre-processed. To create a memory-efficient and useful means of representing the videos, smart video embeddings were created.

\subsubsection{Representing Video Frames}

In our proposed pipeline, the only input to our model is a database of videos. Although videos are rich in information when observed by a human being, computers require alternate methods to process videos. The first step is to analyze videos at the level of their component frames. Once we have extracted frames from the video, we can use image embedding generation techniques to represent them. 

The input videos are sampled at 2 frames per second. Each frame is passed through an image embedding generation model. In our method, we have chosen to use  pre-trained Residual Networks which are trained on the Imagenet dataset to generate frame embeddings. These networks produce embbedings of length 2048. Their residual connections allow features at lower layers to be preserved in deeper layers. We experimented with two variants of the residual network - ResNet50 and ResNet152. These networks are 50 and 152 layers deep respectively. 

\begin{figure}[]
\centerline{\includegraphics[scale=0.6]{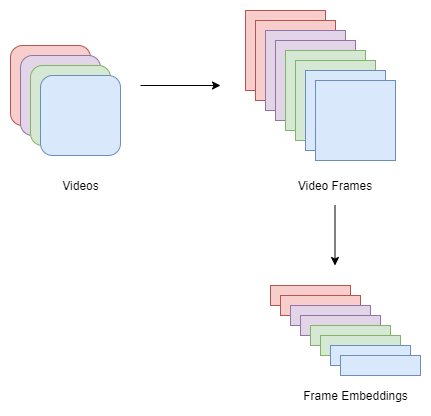}}
\caption{Basic Video Embedding Generation}
\label{fig}
\end{figure}

\subsubsection{Representing Videos}

Videos are generally represented as a concatenation of their component frame embeddings. In our approach, we have chosen to represent the entire dataset of videos together, instead of generating individual video representations. This will allow us to improve the retrieval speed of the model.

Once we have generate embeddings for all sampled frames in the dataset, we cluster them. We use the K-Nearest Neighbors algorithm as a light-weight clustering algorithm that can be used for large-scale clustering applications. Frames across videos in the dataset are assigned into 175 clusters. These clusters are represented by the mean of their component frame embeddings. In this clustering process, we lose out on the important temporal information that is inherently present in videos. To preserve this information, we use a graph-based aggregation technique.

An undirected graph using these clusters is created where each cluster is treated as a node. If frame Y belongs to cluster 1 follows frame X which belongs to cluster 2 in the video, then there is
an edge connecting cluster 1 and 2. The edge weights in the graph are directly proportional to the number of such frame-frame (and cluster-cluster) transitions. To add temporal information to the embeddings, the cluster embeddings are aggregated with their first order neighbor cluster embeddings. This intermediate representation retains the temporal information in the videos and is used in retrieval.

\begin{figure}[]
\centerline{\includegraphics[scale=0.6]{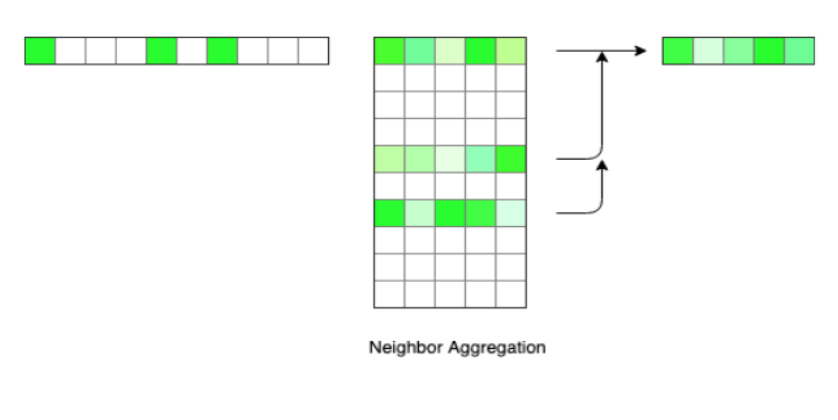}}
\caption{Neighbor Aggregation in Graph Convolutional Networks}
\label{fig}
\end{figure}

\begin{figure}[]
\centerline{\includegraphics[scale=0.4]{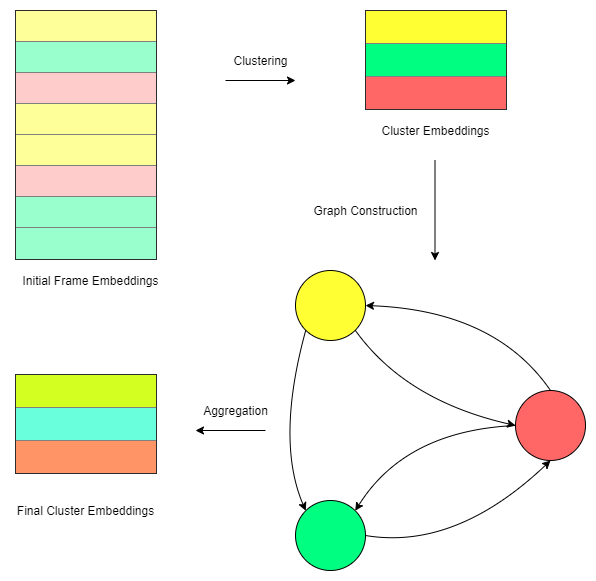}}
\caption{Augmented Embedding Generation} 
\label{fig}
\end{figure}

\subsubsection{Video Retrieval}

Any query image is first processed by the image embedding generation model. We use the augmented cluster embeddings to reduce the search space for every query. The query image embeddings is first compared with the cluster embeddings. The cosine similarity is used as the similarity metric for these embeddings. The clusters are ranked based on their cosine similarities and the top 'c' clusters are chosen for further comparisons. All frame embeddings present in these top clusters are compared with the query image, and ranked based on their similarities. The 'k' number videos corresponding to the top matching frames are retrieved for each query image and Precision@k is calcualted as:
\begin{equation}
    P@k=\frac{R \cap k }{k}
\end{equation}
Where 'R' is the number of videos that are the same category as the query image and 'k' is the total number of videos retrieved. 

After this mAP@k is calculated for all the query images for a particular category. mAP is the mean of all the P@k for all the images for a particular category. It is given as:

\begin{equation}
    mAP=\frac{ \sum_{n=1}^{k} P@k  }{k}
\end{equation}

\section{Dataset}

For the evaluation of this technique experiments were performed on the MSR-VTT dataset. The dataset contains 2990 videos which are around 20-60 seconds long and belong to 20 different categories. This dataset is extensively for video retrieval through text. 

However, rather than using the sentences associated with the videos, this technique uses the video information only hence making it possible to retrieve previously unseen videos.
The categories of videos in the dataset are

1. Music
2. People
3. Gaming
4. Sports, Actions
5. News, Events, Politics
6. Education
7. TV Shows
8. Movie, Comedy
9. Animation
10. Vehicles, Autos
11. How-to
12. Travel
13. Science, Technology
14. Animals, Pets
15. Kids, Family
16. Documentary
17. Food, Drink
18. Cooking
19. Beauty, Fashion
20. Advertisement

For our testing we merged some similar categories like Food and cooking and removed others like movies, documentary, advertisement, etc due to the arbitrary nature of the classes. For example, it's difficult to tell the difference between a movie clip and a clip from  TV show or documentary without any context.

Another reason for excluding some of the other classes like Science and Technology or education, is that there is often no clear visually discernible factor that puts a video in this category. For example a video of a teacher explaining a concept might be classified as education but there's no way to understand that the person in the video is a teacher or that something is being taught.

\begin{figure}[]
\centerline{\includegraphics[scale=0.30]{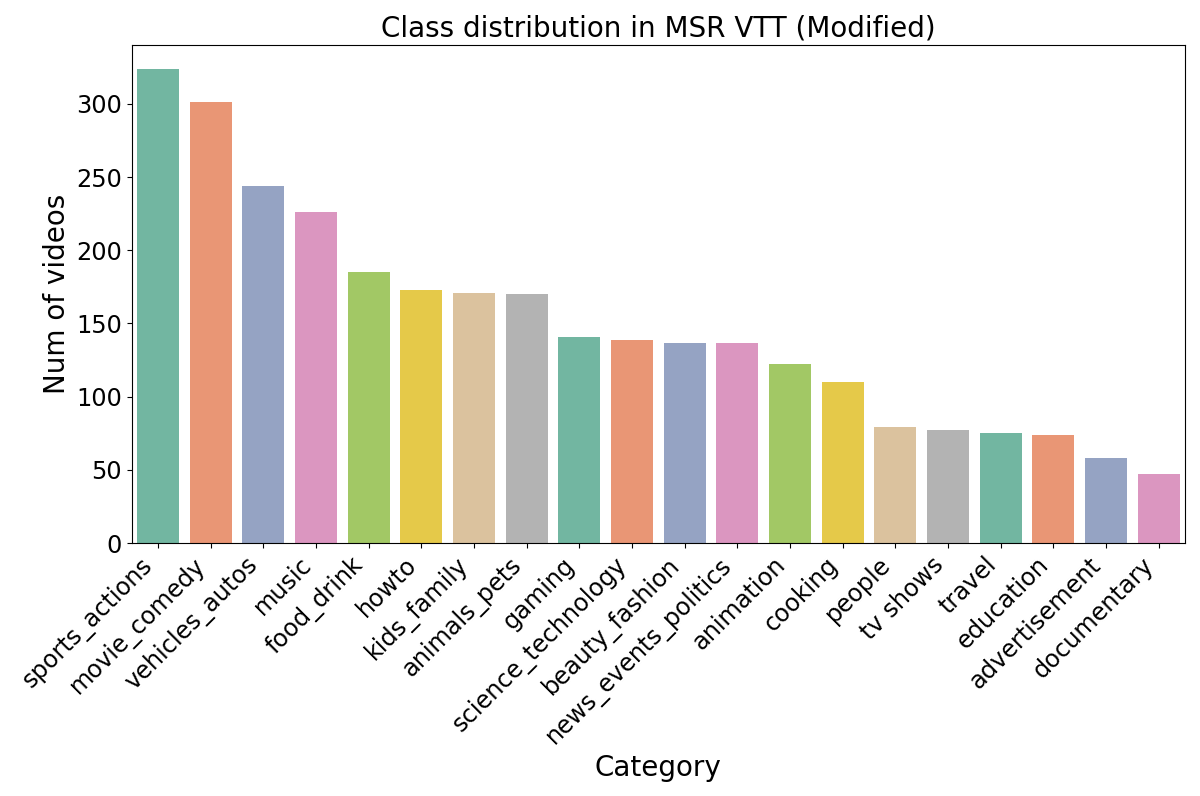}}
\caption{MSR VTT number of videos per class}
\label{fig}
\end{figure}

In the end after filtering, we were left with 11 relevant categories

1. Music
2. Gaming
3. Sports, Actions
4. News, Events, Politics
5. Vehicles, Autos
6. How-to
7. Travel
8. Animals, Pets
9. Kids, Family
10. Food, Drink, Cooking
11. Beauty, Fashion

\begin{figure}[]
\centerline{\includegraphics[scale=0.30]{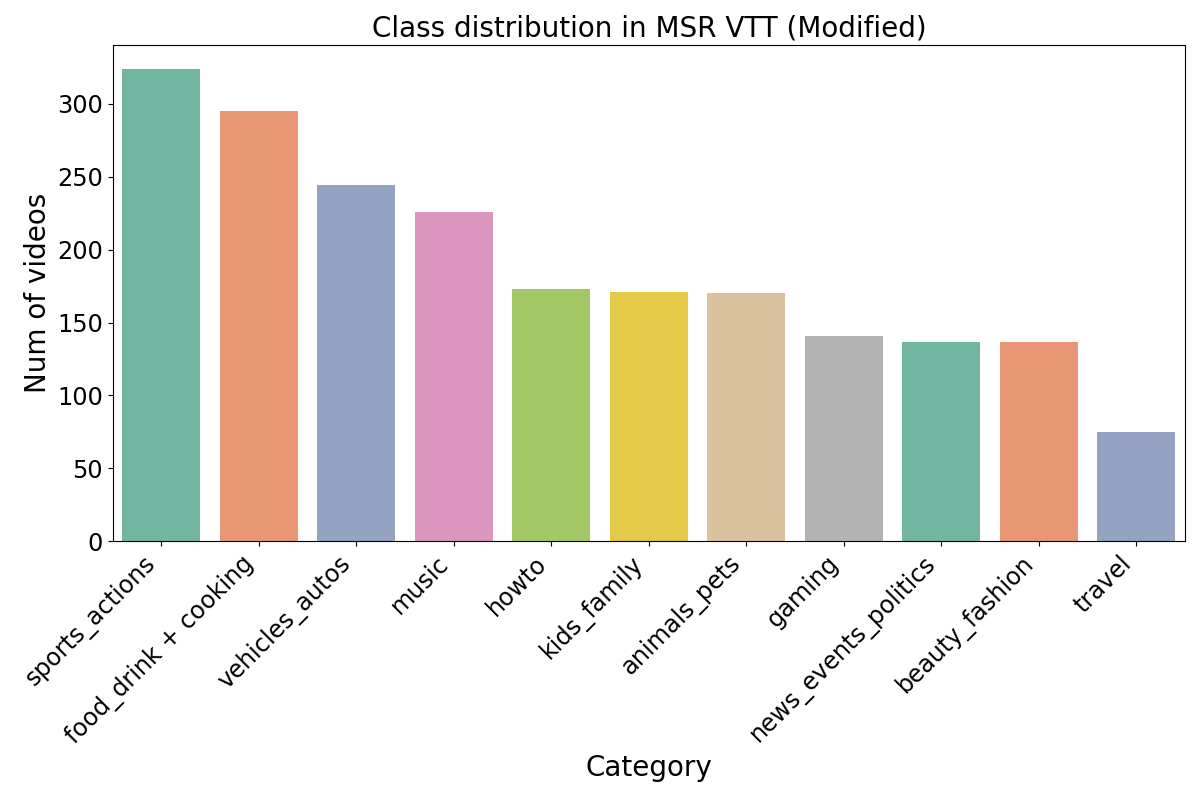}}
\caption{MSR VTT number of videos per class after modification}
\label{fig}
\end{figure}

\begin{figure}[]
\centerline{\includegraphics[scale=0.65]{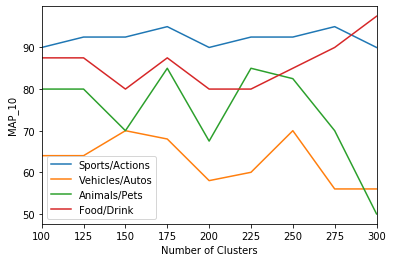}}
\caption{MAP@10 with varying number of clusters}
\label{fig}
\end{figure}

\begin{table}[htbp]
\caption{Model Using ResNet152}
\begin{center}
\begin{tabular}{|c|c|c|c|}
\hline
\textbf{Category} & \textbf{mAP@5} & \textbf{mAP@10} & \textbf{mAP@20} \\
\hline
Music & 60\% & 42.5\% & 37.5\% \\
\hline
Gaming & 50\% & 40\% & 37.5\% \\
\hline
Sports, Actions & 100\% & 97.5\% & 90\% \\
\hline
News, Events, Politics & 45\% & 40\% & 36.25\% \\
\hline
Vehicles, Auto & 85\% & 65\% & 56.25\% \\
\hline
How-to & 30\% & 40\% & 28.75\% \\
\hline
Travel & 55\% & 42.5\% & 32.5\% \\
\hline
Animals, Pets & 85\% & 72.5\% & 61.25\% \\
\hline
Kids, Family & 40\% & 40\% & 32.5\% \\
\hline
Food, Drink, Cooking & 40\% & 40\% & 50\% \\
\hline
Beauty, Fashion & 60\% & 52.5\% & 38.75\% \\
\hline
\end{tabular}
\label{tab1}
\end{center}
\end{table}

\begin{table}[htbp]
\caption{Model Using ResNet50}
\begin{center}
\begin{tabular}{|c|c|c|c|}
\hline
\textbf{Category} & \textbf{mAP@5} & \textbf{mAP@10} & \textbf{mAP@20} \\
\hline
Music & 30\% & 35\% & 31.25\% \\
\hline
Gaming & 55\% & 47.5\% & 36.25\% \\
\hline
Sports, Actions & 100\% & 95\% & 88.75\% \\
\hline
News, Events, Politics & 50\% & 42.5\% & 36.25\% \\
\hline
Vehicles, Auto & 60\% & 57.5\% & 51.25\% \\
\hline
How-to & 40\% & 40\% & 31.25\% \\
\hline
Travel & 50\% & 47.5\% & 35\% \\
\hline
Animals, Pets & 90\% & 70\% & 50\% \\
\hline
Kids, Family & 45\% & 40\% & 33.75\% \\
\hline
Food, Drink, Cooking & 35\% & 37.5\% & 45\% \\
\hline
Beauty, Fashion & 45\% & 45\% & 33.75\% \\
\hline
\end{tabular}
\label{tab1}
\end{center}
\end{table}

\begin{table}[htbp]
\caption{Model Using ResNet152 without creating graph}
\begin{center}
\begin{tabular}{|c|c|c|c|}
\hline
\textbf{Category} & \textbf{mAP@5} & \textbf{mAP@10} & \textbf{mAP@20} \\
\hline
Music & 45\% & 37.5\% & 33.75\% \\
\hline
Gaming & 40\% & 35\% & 38.75\% \\
\hline
Sports, Actions & 100\% & 97.5\% & 91.25\% \\
\hline
News, Events, Politics & 45\% & 47.5\% & 37.5\% \\
\hline
Vehicles, Auto & 80\% & 57.5\% & 50.25\% \\
\hline
How-to & 35\% & 40\% & 31.25\% \\
\hline
Travel & 55\% & 37.5\% & 28.75\% \\
\hline
Animals, Pets & 90\% & 80\% & 61.25\% \\
\hline
Kids, Family & 35\% & 32.5\% & 30\% \\
\hline
Food, Drink, Cooking & 40\% & 42.5\% & 45\% \\
\hline
Beauty, Fashion & 50\% & 42.5\% & 37.5\% \\
\hline
\end{tabular}
\label{tab1}
\end{center}
\end{table}

\begin{table}[htbp]
\caption{Model Using ResNet50 without creating graph}
\begin{center}
\begin{tabular}{|c|c|c|c|}
\hline
\textbf{Category} & \textbf{mAP@5} & \textbf{mAP@10} & \textbf{mAP@20} \\
\hline
Music & 35\% & 32.5\% & 27.5\% \\
\hline
Gaming & 65\% & 55\% & 46.25\% \\
\hline
Sports, Actions & 100\% & 92.5\% & 83.75\% \\
\hline
News, Events, Politics & 50\% & 45\% & 32.5\% \\
\hline
Vehicles, Auto & 68\% & 58\% & 60\% \\
\hline
How-to & 40\% & 41.5\% & 31.25\% \\
\hline
Travel & 40\% & 37.5\% & 30\% \\
\hline
Animals, Pets & 90\% & 72.5\% & 62.5\% \\
\hline
Kids, Family & 35\% & 30\% & 28.75\% \\
\hline
Food, Drink, Cooking & 35\% & 37..5\% & 43.5\% \\
\hline
Beauty, Fashion & 75\% & 52.5\% & 43.75\% \\
\hline
\end{tabular}
\label{tab1}
\end{center}
\end{table}

\begin{table}[htbp]
\caption{Speed Comparison of Models}
\begin{center}
\begin{tabular}{|c|c|}
\hline
\textbf{Model} & \textbf{Effective Search Speed (Video Frames / Second)} \\
\hline
ResNet152 & 15000\\
\hline
ResNet50 & 18000\\
\hline
\end{tabular}
\label{tab1}
\end{center}
\end{table}

\section{Results and Conclusions}
The experiments for this study were run on a system with a 2.2 GHz Intel Core i7 processor with 16 GB of RAM. The system also had an Intel Iris Pro integrated graphics with 1.5 GB of memory.

In the results, P@K denotes the precision of the results in the top ‘K’ ranked videos. The query images
selected to evaluate this model were not from the dataset. Randomly four images for a particular
category were selected and this model was run. For frames in the dataset, the model perfectly
selects the video that it’s from. All the results depicted here are for images not from the dataset.

As seen from the above tables, Resnet-152 outperforms Resnet-50. In a task that is as complex as
this, we expect the larger model to work better than the smaller one. The depth of the ResNet-152 is
more than three times the depth of ResNet-50. This means that there are a lot more weights to train,
and consequently a lot more parameters to learn. However, ResNet-50 searches through the videos
at approximately 180000 frames per second, in comparison to ResNet-152’s 15000 frames per second.

Although this is a considerable difference in speeds, the accuracy of search in ResNet-152 is
significantly higher. Hence, we have chosen to use the ResNet-152 model in the VIRALIQ desktop
application.

Also when the results of the proposed technique which are in tables I and II  are compared to just clustering and retrieving which are given in tables III and IV, we see that the performance of the proposed method is better. The results in table I show improved mAP rates when compared with table III and similarly the results in table II show improved mAP rates when compared with table IV. This is because using this technique we can keep a track of the temporally relevant clusters due to the graph created and retrieve from those clusters as well. However, just by clustering and retrieving, there is a loss in temporal information and hence the precision of the retrieval also falls. In this model the number of temporal clusters to retrieve from can be specified and the best videos from those can be chosen. It can be seen that the results in table I and II outperform the results in table III and IV respectively in almost all categories.

To improve on this further a ResNet model can be pre-trained on a different task, such as scene detection. This would improve the results for certain categories where the composition of the video frames are more important for classification than the individual objects in them. Also this model can be trained on a particular category such as sports, vehicles etc if it's known before hand what domain the model has to work in. Even without this it is evident the proposed model outperforms the traditional model.

\section{Future Work}
The videos in the MSR-VTT dataset are of very short duration. If the videos are longer, another technique can be leveraged to retrieve videos. 

Similar to the proposed technique the videos are sampled at 2 frames per second and then the frames are passed through the chosen residual network to get the embedding of each frame. Each embedding is a vector of
dimension 2048.  The embeddings of each frame are clustered using K-NN. The embedding for each cluster is calculated as the average of all vectors in that cluster. To preserve the temporal information, an undirected graph using these clusters is created as explained previously but here a graph is created for each video.

To also add temporal information to the embeddings as explained, the cluster embedding is also aggregated with its first order neighbour cluster embeddings. Now an incoming image is not compared with individual frames of a video but it is compared with these temporal vectors. The query image is compared to these temporal vectors using cosine similarity. 

The advantage of this technique is that, when  new videos are introduced into the dataset, the embedding and clustering has to be done only for these videos individually. However, in the proposed method the graph might change as the cluster centres will be forced to change due to the addition of new frames. Additionaly, this method produces embeddings for each video and these can be useful in other tasks as well.
This also means that all the frame embeddings don't need to be stored as this method only works with the video embedding. Hence, the overall memory used will be lesser even though the memory access will be the same.
\begin{table}[htbp]
\caption{Results for creating graph for individual videos using ResNet152}
\begin{center}
\begin{tabular}{|c|c|c|c|}
\hline
\textbf{Category} & \textbf{mAP@5} & \textbf{mAP@10} & \textbf{mAP@20} \\
\hline
Music & 50\% & 37.5\% & 33.5\% \\
\hline
Gaming & 45\% & 35\% & 37.5\% \\
\hline
Sports, Actions & 100\% & 95\% & 92.5\% \\
\hline
News, Events, Politics & 50\% & 40\% & 40\% \\
\hline
Vehicles, Auto & 75\% & 57.5\% & 53.75\% \\
\hline
How-to & 30\% & 35\% & 33.75\% \\
\hline
Travel & 55\% & 40\% & 28.75\% \\
\hline
Animals, Pets & 85\% & 77.5\% & 62.5\% \\
\hline
Kids, Family & 35\% & 35\% & 30\% \\
\hline
Food, Drink, Cooking & 40\% & 42.5\% & 46.25\% \\
\hline
Beauty, Fashion & 55\% & 50\% & 40\% \\
\hline
\end{tabular}
\label{tab1}
\end{center}
\end{table}

\begin{table}[htbp]
\caption{Results for retrieval without creating graph using ResNet152}
\begin{center}
\begin{tabular}{|c|c|c|c|}
\hline
\textbf{Category} & \textbf{mAP@5} & \textbf{mAP@10} & \textbf{mAP@20} \\
\hline
Music & 45\% & 37.5\% & 33.5\% \\
\hline
Gaming & 40\% & 35\% & 38.75\% \\
\hline
Sports, Actions & 100\% & 97.5\% & 91.25\% \\
\hline
News, Events, Politics & 45\% & 47.5\% & 37.5\% \\
\hline
Vehicles, Auto & 80\% & 57.5\% & 50\% \\
\hline
How-to & 35\% & 40\% & 31.25\% \\
\hline
Travel & 55\% & 37.5\% & 28.75\% \\
\hline
Animals, Pets & 90\% & 80\% & 61.25\% \\
\hline
Kids, Family & 35\% & 32.5\% & 30\% \\
\hline
Food, Drink, Cooking & 40\% & 42.5\% & 45\% \\
\hline
Beauty, Fashion & 50\% & 42.5\% & 37.5\% \\
\hline
\end{tabular}
\label{tab1}
\end{center}
\end{table}

As seen in tables IV and V, there isn't a big difference between creating a graph or retrieving just after clustering. This can be due to the short nature of the videos in this dataset. Alternatively, different algorithms can be experimented to create the graph to get better results in this technique. The experimentation in this dataset has been limited to only the MSR-VTT dataset as it was the only open source data set which was available in video format.  

\vspace{12pt}

\end{document}